\documentclass[letterpaper, 10 pt, conference]{ieeeconf}  

\IEEEoverridecommandlockouts                              

\makeatletter
\newcommand{\onedot}{.}
\def\@onedot{\ifx\@let@token.\else.\null\fi\xspace}

\def\etal{\emph{et al}\onedot}
\makeatother

\usepackage{etoolbox}
\makeatletter
\patchcmd{\@makecaption}
  {\scshape}
  {}
  {}
  {}
\makeatother

\usepackage{cite}
\usepackage{amsmath,amssymb,amsfonts}
\usepackage{algorithmic}
\usepackage{graphicx}
\usepackage{textcomp}
\usepackage{xcolor}
\def\BibTeX{{\rm B\kern-.05em{\sc i\kern-.025em b}\kern-.08em
    T\kern-.1667em\lower.7ex\hbox{E}\kern-.125emX}}
\usepackage{gensymb}
\usepackage{mathtools}
\usepackage{subfig}
\usepackage{multirow}
\usepackage{siunitx}
\usepackage{rotating}
\usepackage{url}
\usepackage{booktabs}
\usepackage{comment}
\usepackage{adjustbox}
\usepackage{placeins}
\newcommand{\norm}[1]{\left\lVert#1\right\rVert}

\let\labelindent\relax
\usepackage{enumitem}

\usepackage{flushend} 
\usepackage{tikz} 

\makeatletter
\let\NAT@parse\undefined
\makeatother
\usepackage[pagebackref,breaklinks,hidelinks]{hyperref}

\usepackage[capitalize, sort]{cleveref} 
\crefname{section}{Sec.}{Secs.}
\Crefname{section}{Section}{Sections}
\Crefname{table}{Table}{Tables}
\crefname{table}{Tab.}{Tabs.}

\graphicspath{{img/}}

\title{\LARGE \bf
Continual Cross-Dataset Adaptation in Road Surface Classification
}

\author{Paolo Cudrano$^{1}$, Matteo Bellusci$^{1}$, Giuseppe Macino$^{2}$ and Matteo Matteucci$^{1}$
\thanks{$^{1,2}$Department of Electronics Information and Bioengineering, Politecnico di Milano, p.zza Leonardo da Vinci 32, Milan, Italy \hfil {\tt\small name.surname @ \{ polimi.it$^{1}$, mail.polimi.it$^{2}$\}}}
\thanks{This paper is supported by “Sustainable Mobility Center (Centro Nazionale per la Mobilità Sostenibile – CNMS)”  project funded by the European Union NextGenerationEU program within the PNRR, Mission 4 Component 2 Investment 1.4.}
}

\newcommand\copyrighttext{%
\footnotesize \copyright~2023 IEEE. Personal use of this material is permitted. Permission from IEEE must be obtained for all other uses, in any current or future media, including reprinting/republishing this material for advertising or promotional purposes, creating new collective works, for resale or redistribution to servers or lists, or reuse of any copyrighted component of this work in other works.} 
\newcommand\copyrightnotice{%
\begin{tikzpicture}[remember picture,overlay]
\node[anchor=south,yshift=10pt] at (current page.south) {\fbox{\parbox{\dimexpr\textwidth-\fboxsep-\fboxrule\relax}{\copyrighttext}}};
\end{tikzpicture}%
}

\begin{document}
\bstctlcite{IEEEexample:BSTcontrol}

\maketitle

\thispagestyle{empty}
\pagestyle{empty}
\copyrightnotice
%
\begin{abstract}

Accurate road surface classification is crucial for autonomous vehicles (AVs) to optimize driving conditions, enhance safety, and enable advanced road mapping. However, deep learning models for road surface classification suffer from poor generalization when tested on unseen datasets. To update these models with new information, also the original training dataset must be taken into account, in order to avoid catastrophic forgetting. This is, however, inefficient if not impossible, e.g., when the data is collected in streams or large amounts. To overcome this limitation and enable fast and efficient cross-dataset adaptation, we propose to employ continual learning finetuning methods designed to retain past knowledge while adapting to new data, thus effectively avoiding forgetting. Experimental results demonstrate the superiority of this approach over naive finetuning, achieving performance close to fresh retraining. While solving this known problem, we also provide a general description of how the same technique can be adopted in other AV scenarios. We highlight the potential computational and economic benefits that a continual-based adaptation can bring to the AV industry, while also reducing greenhouse emissions due to unnecessary joint retraining.

\end{abstract}


\section{INTRODUCTION}
\label{sec:introduction}

Autonomous vehicles (AVs) have gained significant attention in recent years, with tremendous advancements in various aspects of their operation. One crucial aspect is the detection and understanding of the road environment, which includes the ability to classify different types of road surfaces. Road surface classification is critical for AVs as it provides valuable information about the road material, allowing the vehicle to adapt its driving operating conditions accordingly.
For instance, when traversing an unpaved rural road, the vehicle can modulate its velocity and adjust its steering to navigate safely through the challenging terrain. Similarly, on a cobblestone road, the presence of rain can increase the risk of skidding, necessitating proactive measures to avoid potential hazards. In addition, road surface classification facilitates the creation of high-definition (HD) maps, enabling AVs to receive information about road conditions in advance. With this information, vehicles can plan their routes more effectively, enhancing the overall driving experience. Moreover, knowledge of the pavement material is crucial when developing realistic simulations for testing and validating AVs and Advanced Driver Assistance Systems (ADAS). Indeed, this knowledge enables the recreation of the driving scenario with higher fidelity.

\begin{figure*}[t]%
    \centering%
    \includegraphics[width=\linewidth]{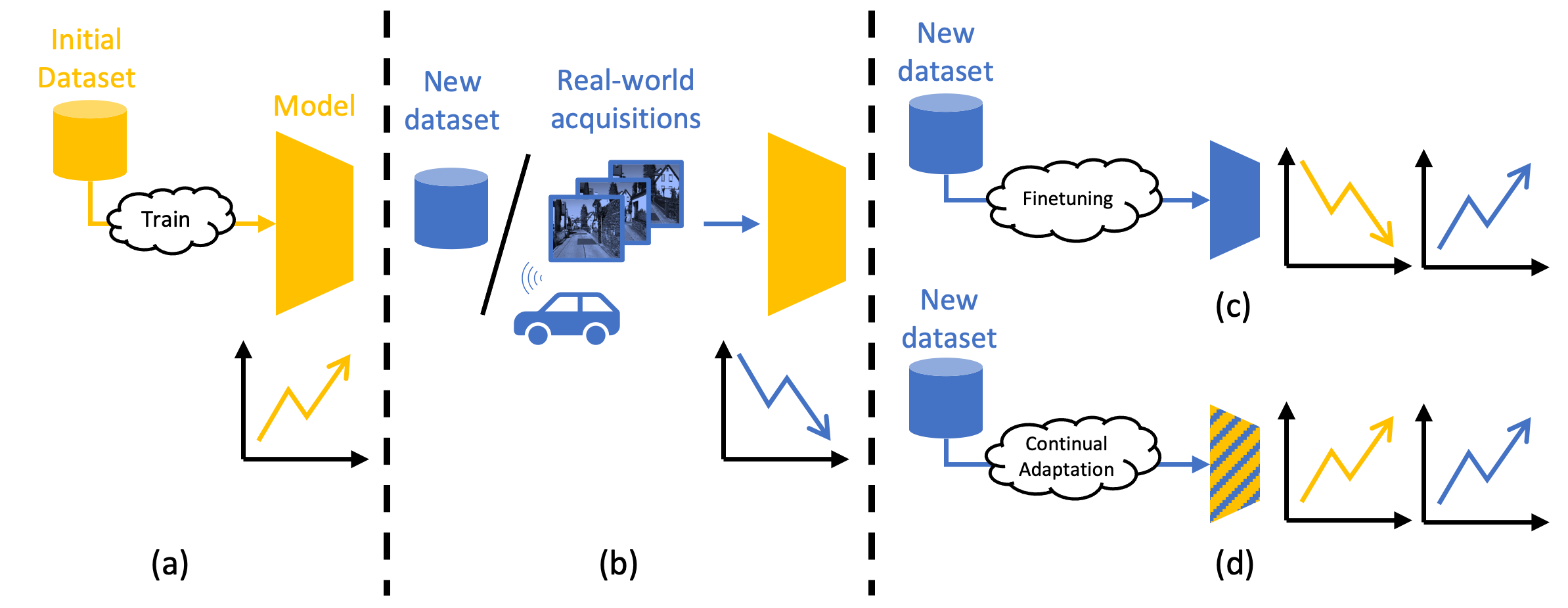}%
    \caption{Models adopted in AVs are typically trained on a single well-known dataset~(yellow). These models seem to perform well on the test data from the same dataset~(a), but when applied to other datasets or in the real-world~(blue), their performance drops as they lack generalization~(b). It is crucial to still update such models as new data is available. Finetuning only on new data, however, leads to catastrophic forgetting, where the model learns the new information but loses previous knowledge~(c). Instead, we propose to adopt continual learning strategies to allow for model updates while retaining past knowledge~(d).}%
    \label{fig:intro}%
\end{figure*}

To tackle the road surface classification problem, current approaches rely on deep learning techniques due to their exceptional ability to extract intricate features from visual data. However, deep learning models heavily rely on the quality and diversity of their training dataset, and several issues can arise when applying these models to real-world scenarios~\cite{tommasi2017deeper}. 
For instance, biased or unrepresentative training data can lead to models that perform poorly or unfairly~\cite{pessach2022review}. Additionally, the lack of diversity in the training data can limit the ability of models to generalize to unseen situations or novel datasets.  
It is known that the distribution shift between different datasets can indeed significantly impact model performance, as models trained on one dataset often fail to generalize when tested on other datasets (\cref{fig:intro}a\nobreakdash--b), i.e., they show poor cross-dataset generalization~\cite{torralba2011unbiased}.

In the context of AVs, a large number of datasets have been collected. For the task of road surface classification, in particular, three datasets are publicly available and contain all the necessary information: RTK~\cite{rateke_road_2019}, KITTI~\cite{Geiger2012CVPR}, and CaRINA~\cite{7795529}. It has been shown in the literature~\cite{rateke_road_2019} that existing deep learning models trained on one of these three datasets exhibit inadequate generalization when tested on the remaining two. This limitation poses a significant obstacle to the deployment of robust road surface classification systems in autonomous vehicles, as it highlights how these models cannot operate reliably in all plausible real-world scenarios.

Addressing this cross-dataset generalization problem is critical. A straightforward approach would be to perform finetuning on the model using the target dataset. However, finetuning only on novel data leads to catastrophic forgetting~\cite{french1999catastrophic}, i.e., the model's performance on previously learned tasks quickly deteriorates as it becomes optimized only for the new data (\cref{fig:intro}c). A consequent idea would be to include the original dataset in the data used for finetuning, forcing the model to remember past data while learning from novel samples. Including the entire original dataset in the training, however, can be computationally inefficient, especially when dealing with large historical datasets.
Additionally, access to past datasets may not even be feasible in settings of online training, or it may be forbidden due to privacy restrictions.

To overcome this limitation, an alternative solution is necessary: updating the model solely based on the new data, yet avoiding forgetting past knowledge (\cref{fig:intro}d). 
This problem is precisely the focus of continual learning (CL), a research field studying how to mitigate catastrophic forgetting.

In this work, we exploit one such continual learning strategy to solve the problem of dataset adaptation in road surface classification.
In doing so, our contribution is twofold. First, we propose a practical solution to the known problem of poor cross-dataset generalization in road surface classification, showing how it is possible to efficiently update a model using only new data. Second, we showcase the applicability of continual learning strategies not only to this specific task, but to any AV task that necessitates updating a model with new data effectively yet efficiently.

\section{RELATED WORK}

Road surface classification has been addressed by several works in the literature.
Pereira et~al.~\cite{pereira2018classification} distinguish between paved and unpaved road surfaces using a Convolutional Neural Network (CNN) based on VGG~\cite{simonyan2014very}. They train their deep model with smartphone-collected images acquired under several weather conditions.
Nolte et~al.~\cite{nolte2018assessment}, instead, compare CNN architectures at classifying the road surface (asphalt, dirt, grass, wet, cobblestone, snow) under different weather conditions, exploiting various datasets as Oxford RoboCar~\cite{RobotCarDatasetIJRR} and KITTI~\cite{Geiger2012CVPR}.
Using a custom CNN architecture, Rateke \etal~\cite{rateke_road_2019} classify the road material (asphalt, paved, unpaved), and then proceed to analyze road quality and conditions (e.g., presence of potholes) with an ad-hoc system.
They train their road material classifier using monocular images from KITTI~\cite{Geiger2012CVPR}, CaRINA~\cite{7795529}, and to properly train on unpaved conditions, they additionally propose a new purposefully-collected dataset, RTK. In the same work, Rateke \etal~\cite{rateke_road_2019} also report cross-dataset generalization issues when training on one single dataset and testing on the other two. They propose to mitigate this problem by training their model jointly on all three datasets.
This, however, is not only inefficient, but it implies that, if a criticality is discovered in the future, retraining on all available datasets will be necessary again in order to fix it.

Similar poor cross-generalization performances are reported by Gesnouin \etal~\cite{gesnouin_assessing_2022} in the context of pedestrian crossing detection, a completely different task in autonomous vehicle perception. This shows that the problem of poor cross-dataset generalization is widespread and hinders the reliability of potentially any AV perception system. A systematic approach must be found to solve it consistently.

As continual learning gains attention in the deep learning community, applications in industrial and real-world setups are emerging, including works on other autonomous driving problems.
Shaheen \etal~\cite{shaheen_continual_2022} propose a review of the literature and discuss emerging applications and related challenges for autonomous vehicles, including insights on scene recognition, lane keeping, and Reinforcement Learning (RL)-based vehicle control. 
Verwimp \etal~\cite{verwimp_clad_2022}, instead, recently proposed a new benchmark for continual learning, specifically designed for autonomous vehicle tasks.
Zhang and Mueller~\cite{zhang_claire_2022} assess the feasibility of online continual learning for on-board vehicle devices, proposing a lightweight object detection architecture.
Kalb \etal~\cite{kalb_continual_2021} adapt the formulation of semantic segmentation to a continual learning setup, validating their work on autonomous driving benchmarks. In a successive work, Kalb \etal~\cite{kalb_improving_2022} further investigate the performance of replay techniques when dealing with segmentation tasks, field previously almost unexplored.
In segmentation, Termöhlen \etal~\cite{termohlen_continual_2021} use frequency-based style transfer to perform continual domain adaptation and contrast catastrophic forgetting.
Also traffic monitoring and forecasting has seen interest in early continual learning works. Among these, Tsai \etal~\cite{tsai_incremental_2022} perform traffic forecasting using an online version of EWC~\cite{kirkpatrick2017overcoming}, with application to traffic pattern discovery during COVID-19 road regulation changes. Sun \etal~\cite{sun_self-supervised_2022}, instead, apply continual techniques to a camera-based traffic advisory system.

\section{CONTINUAL LEARNING}

Before framing the problem of road surface classification in the context of continual learning, we briefly present the field and its current developments.

Continual Learning (CL) aims at making learning systems able to acquire knowledge through time from a possibly-infinite stream of data, without forgetting past learned information.
The main problem faced by CL is catastrophic forgetting, i.e., the inability of deep learning models to consistently 
retain past knowledge when trained on new data.

This characteristic is rooted in the optimization procedure adopted to train these models.
Let us consider, without loss of generality, the most simple scenario of a deep learning model (or network), whose parameters (or weights) are optimized (or trained) in a fully supervised setting on real-world high-dimensional data, e.g., camera images. Let us also assume we have a mechanism to sample input data and associated target outputs from their real-world statistical distribution.

To find the best weights for our model, we start acquiring a random data sample and its associated target output, and use our model with its current weights to predict an output. We can then compare this prediction to its associated target by means of a loss function (e.g., a distance measure), which assesses the model's performance. As model and loss function are smooth, by chain rule we can compute the gradient of this loss with respect to each model weight, and perform a step of gradient descent (or variations) in such direction. Doing so will slightly pull the weights of our model towards a configuration that performs better on the current data sample.

Repeating this procedure over different samples, the weight updates give rise to a tug-of-war in the weight space, pulling weights in the optimal direction for each data instance. We can expect that, after a large number of samples, the weights converge to a region of compromise that performs adequately over all data samples.
It can be shown that this indeed happens, but only if our sampling procedure generates \textit{i.i.d.} (independent and identically distributed) samples over the \textit{entire data distribution}. Intuitively, this means that the samples must span over the entire distribution and that regions of such distribution are eventually revisited.

Clearly, this condition is strict and unrealistic for real-world data, but to approach it, two steps are always taken. First, a very large amount of samples is used in the optimization, in order to best approximate the entire data distribution; and second, these samples are shuffled, in order to break any inherent dependency.
To carry on these two operations practically, these samples are preemptively collected and stored in the form of a dataset.
It is for this reason that, trying to run this procedure on an already-trained model and sampling only from a new dataset, inevitably leads to catastrophic forgetting, as we are sampling only from a portion of the entire data distribution of interest, thus violating the above condition.
Intuitively, this is because the tug-of-war mechanism internal to the optimization is not balanced, and will end up favoring the performance of the model only on newly-seen data, at the expense of older ones.

Although solutions for catastrophic forgetting are still currently under research, several training strategies have been proposed in the literature to mitigate it, finding a trade-off for the stability-plasticity dilemma~\cite{lesort2020continual}. Such strategies are traditionally categorized into three classes: replay-based, architecture-based, and regularization-based~\cite{lesort2020continual}. In recent times, hybrid versions~\cite{pellegrini2020latent} are also emerging.

\textbf{Replay-based strategies}, such as iCaRL~\cite{rebuffi2017icarl}, DGR~\cite{shin2017continual}, and GEM~\cite{lopez2017gradient}, 
represent the most straightforward approaches. They focus on storing in a memory buffer a few of the seen samples during training, in order to mix them with the new data when learning future experiences. In this way, they recreate a simile-i.i.d. setup and thus maintain good performance on past experiences. The main focus in this direction is devoted, of course, to the selection of as few most representative samples as possible, in order to maintain low memory consumption. Generative models and latent replay are also commonly adopted as rehearsal strategies~\cite{de2021continual}.

\textbf{Architecture-based strategies}, instead, physically modify the network architecture to account for the incoming new information. 
A simple baseline consists in adding a new head to the network every time a new experience is encountered. This expedient untangles stability and plasticity, as it learns custom weights for each experience, at the cost of rapidly-increasing memory consumption. 
Among others, examples are progressive neural networks (PNN)~\cite{rusu2016progressive} and HAT~\cite{serra2018overcoming}.

\textbf{Regularization-based strategies}, at last, deal with catastrophic forgetting by constraining the weight updates through regularized losses, in order to prevent large changes of the weights more responsible for modeling past data distributions. 
In a sense, these techniques deal with the root of catastrophic forgetting, controlling the tug-of-war mechanism and dealing with the problem of credit assignment~\cite{hadsell2020embracing}.
Based on theoretical neuroscience considerations on synaptic consolidation~\cite{parisi2019continual, hadsell2020embracing}, these approaches provide the advantage of having low extra-memory requirements.
Among these methods, Elastic Weight Consolidation (EWC)~\cite{kirkpatrick2017overcoming} limits weights' gradients proportionally to their importance for previous experiences. These importances are estimated by computing their Fisher information matrix at the end of each experience. Synaptic Intelligence (SI)~\cite{zenke2017continual}, instead, estimates similar weight importances in an online manner, allowing for a smooth online transition between consecutive experiences. 
These strategies, however, must be applied from the beginning of the training, and thus cannot be adopted with pretrained models. 
Conversely, Less-Forgetful Learning (LFL)~\cite{jung2018less} directly uses the new data as a proxy for the old ones. In this way, it is able to regularize the embedding space generated by the last layer of the classifier.

\section{CONTINUAL ADAPTATION TO NEW DATA}
\label{sec:continual_adaptation}

With a background in CL, we can now frame the problem of dataset adaptation in the context of continual learning. Specifically, we focus our attention on the task of road surface classification, although the following formulation is general to other AV applications.

In road surface classification, we consider a deep network $F_\theta$ of parameters $\theta$ that has the task of classifying the road surface between three categories: asphalt, unpaved (e.g.,~gravel roads) and paved (other than asphalt, e.g.,~cobblestone).
Let us assume that $F$ is first trained on dataset $D_0$. We can denote this trained network with $F_\theta = F_{\theta_0^*}$, where $\theta_0^*$ represents the optimal parameters obtained training on $D_0$. Notice that $D_0$ could be a large pretraining dataset and might be unavailable at a later time.
Our objective is to adapt the model weights $\theta$ to a new dataset $D_i$ such that, while new knowledge is acquired, the network retains high performance also on all previously seen datasets ($D_0, \dots, D_{i-1}$). Here, we use $i$ to index experiences, which correspond with a new dataset being presented to the network.

In a general continual learning setup, at every new experience, changes can occur in the distribution of the input data, the distribution of the targets, or the set of target labels. In dataset adaptation, however, we consider only changes in the input data distribution. 
This setup is also known in the literature as domain-incremental learning~\cite{de2021continual}. Nevertheless, our formulation can be adapted for task-incremental, class-incremental, and data-incremental setups~\cite{de2021continual}.

Among all possible strategies proposed in the literature, for the task of dataset adaptation we rely on Less-Forgetful Learning (LFL), proposed by Jung \etal~\cite{jung2018less}. In the following, we summarize its working mechanisms and highlight its main advantages over the alternatives.

\subsection{Less-Forgetful Learning (LFL)}
\label{sec:lfl}
Less-Forgetful Learning (LFL)~\cite{jung2018less} 
is a regularization-based strategy perfect for the problem of continual domain adaptation, i.e., avoiding catastrophic forgetting due to input domain shifts.
LFL can be applied on pretrained models as-is, as it does not require any additional computation during training on the first dataset.
This is a crucial feature in the context of this work, as we aim to rapidly adapt already existing models as new information becomes available.

Considering a deep feed-forward neural network with a final softmax classification layer, the authors of LFL consider the top layer $F^L$ as a linear classifier, while the rest of the network $F^{L-1}$ is considered as a feature extractor, mapping the input data to a linearly separable embedding space. Building on this consideration, the authors then highlight two key desiderata for their continual strategy:
\begin{enumerate}[label=\Alph*.]
    \item \label{enum:lfl_desiderata1} The decision boundaries outlined by the top layer $F^L$ must remain unchanged between different experiences.
    \item \label{enum:lfl_desiderata2} The embedding of data from past experiences should not move significantly when updating the network.
\end{enumerate} 
Clearly, \ref{enum:lfl_desiderata2} cannot be guaranteed in general, as it requires constant access to past datasets. Nevertheless, the authors find that using data from new experiences as a proxy for the old ones still provides good performances. As a consequence, given a model $F_\theta=F_{\theta_0^*}$, pretrained on datasets $D_0$, their strategy for updating $F_\theta$ using dataset $D_1$ prescribes to:
\begin{enumerate}[wide, labelwidth=!, labelindent=0pt]
    \item Copy the current optimal network weights $\theta_1 \coloneqq \theta_{0}^*$.
    \item \label{enum:lfl_alg2} Freeze the top softmax layer $F^L$ (to achieve \ref{enum:lfl_desiderata1}).
    \item Train $F_{\theta}$ on dataset $D_1$ with loss:
    \begin{equation}
        \mathcal{L}(x; \theta_1, \theta_0^*) = \mathcal{L}_c(x; \theta_1) + \lambda_e \mathcal{L}_e(x; \theta_1, \theta_0^*) + \mathcal{R}(\theta_1), 
    \end{equation}
    where $x \in D_1$, $\mathcal{R}$ is a regularization term (such as an $L_2$ penalty), and $\mathcal{L}_e$ (enforcing \ref{enum:lfl_desiderata2}) is:
    \begin{equation}
        \mathcal{L}_e(x; \theta_1, \theta_0^*) = \frac{1}{2} \norm{ F_{\theta_0^*}^{L-1}(x) - F_{\theta_1}^{L-1}(x) }_2.
    \end{equation}
\end{enumerate}
This procedure can be repeated for any further new dataset $D_i$ becoming available over time. In the end, $F_\theta$ will have converged to a configuration that has acquired as much knowledge as possible from the last dataset, without forgetting what was learned from past datasets.

\section{EXPERIMENTS}

\subsection{Experimental setup}

To validate the efficacy of our methodology in achieving efficient dataset adaptation, we follow the same experimental setup proposed by Rateke \etal~\cite{rateke_road_2019}. In particular, we replicate their findings that the task of road surface classification presents poor cross-dataset generalization, and we show how the continual adaptation strategy presented in \cref{sec:continual_adaptation} significantly mitigates the problem without needing to store and process past data.

\begin{figure}[t]%
    \centering%
    \includegraphics[width=\columnwidth]{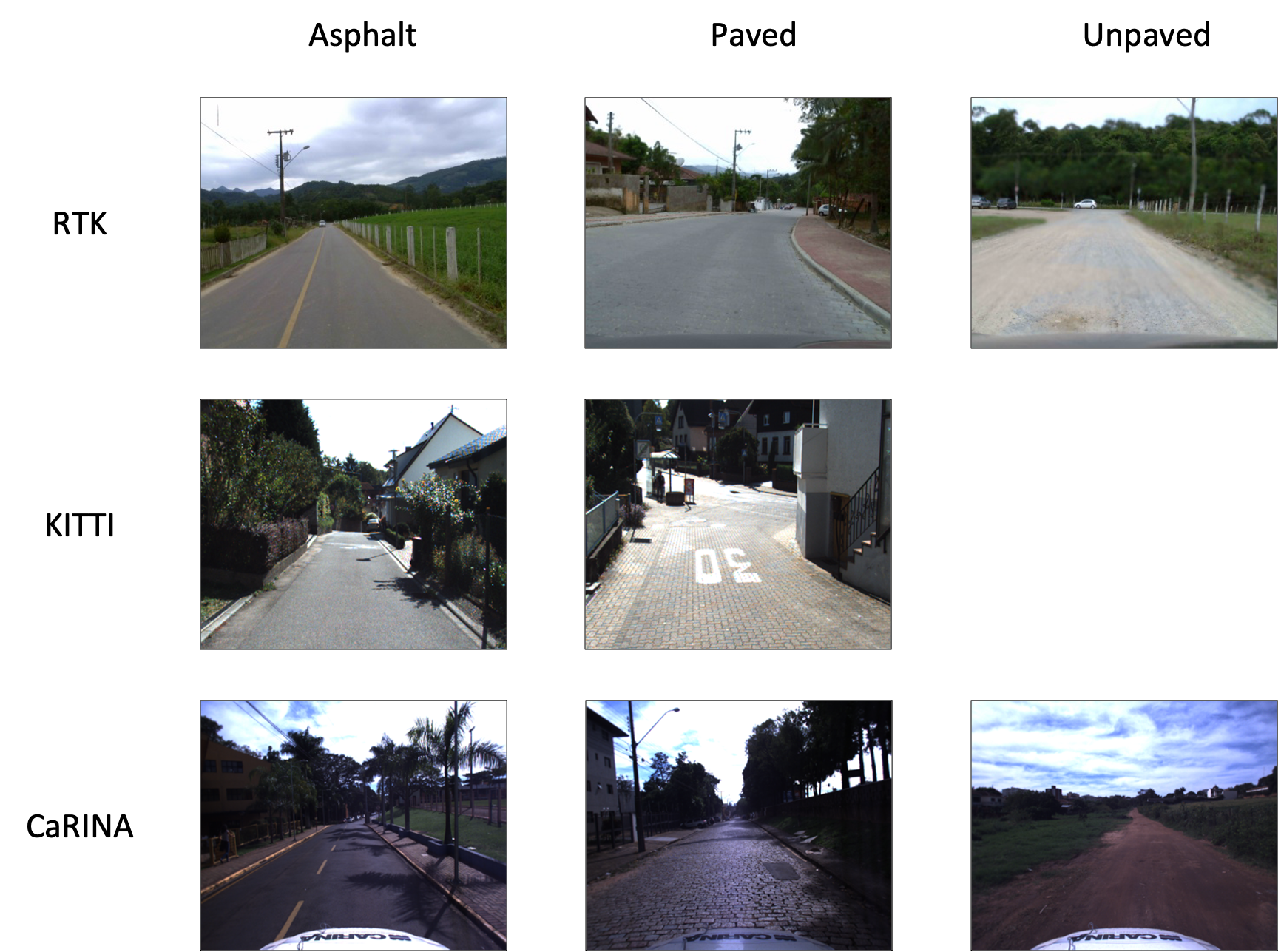}%
    \caption{The three mainly known public datasets for road surface classification: RTK, KITTI, and CaRINA. They provide three label categories: asphalt, paved (e.g., cobblestone), and unpaved. Empty spaces are left where no data is present.}%
    \label{fig:dataset}%
\end{figure}

To maintain replicability, we employ the CNN architecture described in their work~\cite{rateke_road_2019}.
Moreover, we employ the same three datasets: RTK~\cite{rateke_road_2019}, KITTI~\cite{Geiger2012CVPR}, and CaRINA~\cite{7795529}. Each dataset is divided into train and test splits. \cref{fig:dataset} displays a few samples of the images present in each dataset, highlighting the three possible output labels: asphalt, unpaved, and paved.
In their work, \cite{rateke_road_2019} consider RTK as the first training source ($D_0$ in our notation) because of its larger size, more suitable for the initial training; KITTI and CaRINA instead, containing fewer samples, are considered as secondary datasets ($D_1, D_2$).
When training only on RTK and testing on KITTI and CaRINA, Rateke \etal\ report significantly lower metrics on the latter two, highlighting poor cross-dataset generalization. For this reason, they then repeat the experiment training jointly on all three datasets, achieving satisfactory performances on all three test sets, at the cost of a longer and more expensive training.
We replicate their results and extend their experimental setup to evaluate the performance of the reference model when trained using two baselines (naive and joint training~\cite{de2021continual}) and the LFL continual learning strategy. In particular:

\begin{table}[t!]
\renewcommand{\arraystretch}{1.25}
\vspace{.5em}
\caption{Naive strategy}
\label{tab:res_naive}
\begin{adjustbox}{width=\linewidth,center}
\begin{tabular}{@{}l|rr|rr|rr@{}}
\toprule
     & \multicolumn{6}{c}{Testing dataset}\\
     & \multicolumn{2}{c|}{RTK} & \multicolumn{2}{c|}{KITTI} & \multicolumn{2}{c}{CaRINA}\\
    Training dataset & AUROC & F1 & AUROC & F1 & AUROC & F1 \\
\hline
RTK                     & 0.9932 & 0.9614 & 0.3778 & 0.4462 & 0.7206 & 0.5660\\
RTK $\to$ KITTI             & 0.7526 & 0.5083 & 0.6521 & 0.9462 & 0.4564 & 0.3868\\
RTK $\to$ KITTI $\to$ CaRINA    & 0.5972 & 0.6924 & 0.6189 & 0.8385 & 0.9995 & 0.9717\\
\bottomrule
\end{tabular}
\end{adjustbox}
\end{table}
\begin{table}[t!]
\renewcommand{\arraystretch}{1.25}
\caption{LFL strategy}
\label{tab:res_lfl}
\begin{adjustbox}{width=\linewidth,center}
\begin{tabular}{@{}l|rr|rr|rr@{}}
\toprule
     & \multicolumn{6}{c}{Testing dataset}\\
     & \multicolumn{2}{c|}{RTK} & \multicolumn{2}{c|}{KITTI} & \multicolumn{2}{c}{CaRINA}\\
    Training dataset & AUROC & F1 & AUROC & F1 & AUROC & F1 \\
\hline
RTK                     & 0.9932 & 0.9614 & 0.3778 & 0.4462 & 0.7206 & 0.5660\\
RTK $\to$ KITTI             & 0.9742 & 0.9186 & 0.6278 & 0.8692 & 0.6606 & 0.5660\\
RTK $\to$ KITTI $\to$ CaRINA    & 0.9774 & 0.9338 & 0.6115 & 0.8077 & 0.9983 & 0.9528\\
\bottomrule
\end{tabular}
\end{adjustbox}
\end{table}
\begin{table}[t!]
\renewcommand{\arraystretch}{1.25}
\caption{Joint strategy}
\label{tab:res_joint}
\begin{adjustbox}{width=\linewidth,center}
\begin{tabular}{@{}l|rr|rr|rr@{}}
\toprule
     & \multicolumn{6}{c}{Testing dataset}\\
     & \multicolumn{2}{c|}{RTK} & \multicolumn{2}{c|}{KITTI} & \multicolumn{2}{c}{CaRINA}\\
    Training dataset & AUROC & F1 & AUROC & F1 & AUROC & F1 \\
\hline
RTK $\cup$ KITTI $\cup$ CaRINA    & 0.9852 & 0.9648 & 0.6353 & 0.8769 & 0.9848 & 0.9340\\
\bottomrule
\end{tabular}
\end{adjustbox}
\end{table}
\begin{table}[t!]
\renewcommand{\arraystretch}{1.25}
\captionsetup{justification=centering}
\caption{Comparison of the final model\\against the Joint strategy}
\label{tab:res_comparison}
\begin{adjustbox}{width=\linewidth,center}
\begin{tabular}{@{}l|rr|rr|rr@{}}
\toprule
     & \multicolumn{6}{c}{Testing dataset}\\
     & \multicolumn{2}{c|}{RTK} & \multicolumn{2}{c|}{KITTI} & \multicolumn{2}{c}{CaRINA}\\
    Strategy & $\Delta$AUROC & $\Delta$F1 & $\Delta$AUROC & $\Delta$F1 & $\Delta$AUROC & $\Delta$F1\\
\hline
Naive  & 0.3880 & 0.2724 & 0.0164 & 0.0384 & $-$0.0147 & $-$0.0377\\ 
LFL    & 0.0078 & 0.0310 & 0.0238 & 0.0692 & $-$0.0135 & $-$0.0188\\
\bottomrule
\end{tabular}
\end{adjustbox}
\vspace{-.5em}
\end{table}

\begin{description}
    \item[Naive finetuning.] After our model has been trained on RTK, we continue finetuning it, first on KITTI and then on CaRINA. This replicates the findings of~\cite{rateke_road_2019} and represents a lower bound for our expected performance, as it is the best setup for catastrophic forgetting to occur.
    \item[LFL (continual adaptation).] After our model has been trained on RTK, we use the continual learning strategy LFL (Sec.~\ref{sec:lfl}) to finetune it first on KITTI and then on CaRINA. Notice that LFL does not need to be applied when training on RTK.
    \item[Joint training.] We discard our model trained only on RTK, and retrain it from scratch over the combination of all three datasets, as proposed in~\cite{rateke_road_2019}. This represents an upper bound to our expected performance as, merging and shuffling all available data, no forgetting can occur.
\end{description}

Our entire experimental setup is implemented in PyTorch~\cite{paszke2019pytorch}, using the Avalanche framework~\cite{lomonaco2021avalanche}. We train each experience for 30 epochs, using an SGD optimizer with learning rate of $0.002$ and $\lambda_e = 1$. These values have been finetuned to yield the best results and thus propose a fair evaluation. All experiments are executed on an Nvidia Quadro RTX 6000 GPU.

\subsection{Experimental results}

We evaluate the performance of the reference model in terms of AUROC and F1-score, with the former being the most indicative as it balances precision and recall and is not affected by class imbalance.

To evaluate the impact of forgetting, we compute these metrics after training on the initial dataset (RTK), and after adding each additional dataset (KITTI, CaRINA). These metrics are computed individually on the test splits of each dataset.
In this way, we can also evaluate the performances yielded on KITTI and CaRINA even before they are seen by the model, showing the model's cross-dataset performance.

\Cref{tab:res_naive,tab:res_lfl,tab:res_joint} report the results of this analysis for Naive, LFL, and Joint strategies respectively.
Notice that the Naive strategy can be considered as an approximate lower bound for performances, while the Joint strategy as an approximate upper bound.

From \Cref{tab:res_naive,tab:res_lfl}, we notice that both Naive and LFL strategies report the same performance when training on the first dataset, RTK. This is because LFL does not intervene during the first training, thus both strategies perform the same exact computation at this stage. 
We notice that while the model has great performance on RTK, it generalizes poorly to both unseen datasets KITTI and CaRINA, despite the classification task being the same. This replicates the poor cross-dataset generalization reported by Rateke \etal~\cite{rateke_road_2019}.

When finetuning on the second dataset (KITTI), Naive and LFL present different behaviors. In both cases, the performance on KITTI rises, as the model is adapting to its data distribution. However, Naive leads to a significant drop in performance on the RTK test set, going from an AUROC of 0.9932 to 0.7526. This is a clear manifestation of catastrophic forgetting: the model has adapted to the new data distribution but has lost performance on the previous distribution. 
LFL, on the contrary, displays only a minor perturbation, and its AUROC performance on RTK remains above 0.97 even when learning from a new dataset.
The same trend is noticeable when we move to the third dataset (CaRINA), as the AUROC on RTK drops further to 0.5972 under the Naive strategy. This confirms that this strategy leads to significant catastrophic forgetting, while LFL maintains a high performance throughout the model's lifetime.

When comparing to the upper bound performance given by the Joint strategy in \Cref{tab:res_comparison}, we notice that the final model obtained through sequential updates achieves comparable performance when adopting the LFL strategy. Indeed, the model obtained with LFL reports a difference of at most 2.38\% AUROC on any dataset. The Naive strategy, instead, leads to drops in AUROC of 38.80\% on RTK, significantly impacting the reliability of the system on past data.

As current CL strategies are not perfectly mature, we realize that the performance of LFL remains slightly inferior to a Joint training (\Cref{tab:res_comparison}). However, if datasets related to past experiences are not available, this is the best we can achieve, and it improves significantly over the Naive approach.
Moreover, even if we do have access to all past data, we might still want to consider whether to use it entirely. Indeed, we must also take into account the speedup induced by continual adaptation. In our experiments, finetuning only on the last dataset was found to be 5 times faster than training using all past data.
In this work, the dataset dimensions are modest, and thus the computation required could still be handled. However, with large-scale datasets, the saved time is magnified significantly.
As a result, when dealing with similar problems of dataset adaptation, we advise practitioners to evaluate the trade-off between the slight performance loss experienced by continual adaptation and the efficiency gained in terms of computation resources.

\begin{figure}[t!]%
    \centering%
    \subfloat{%
        \includegraphics[width=\columnwidth]{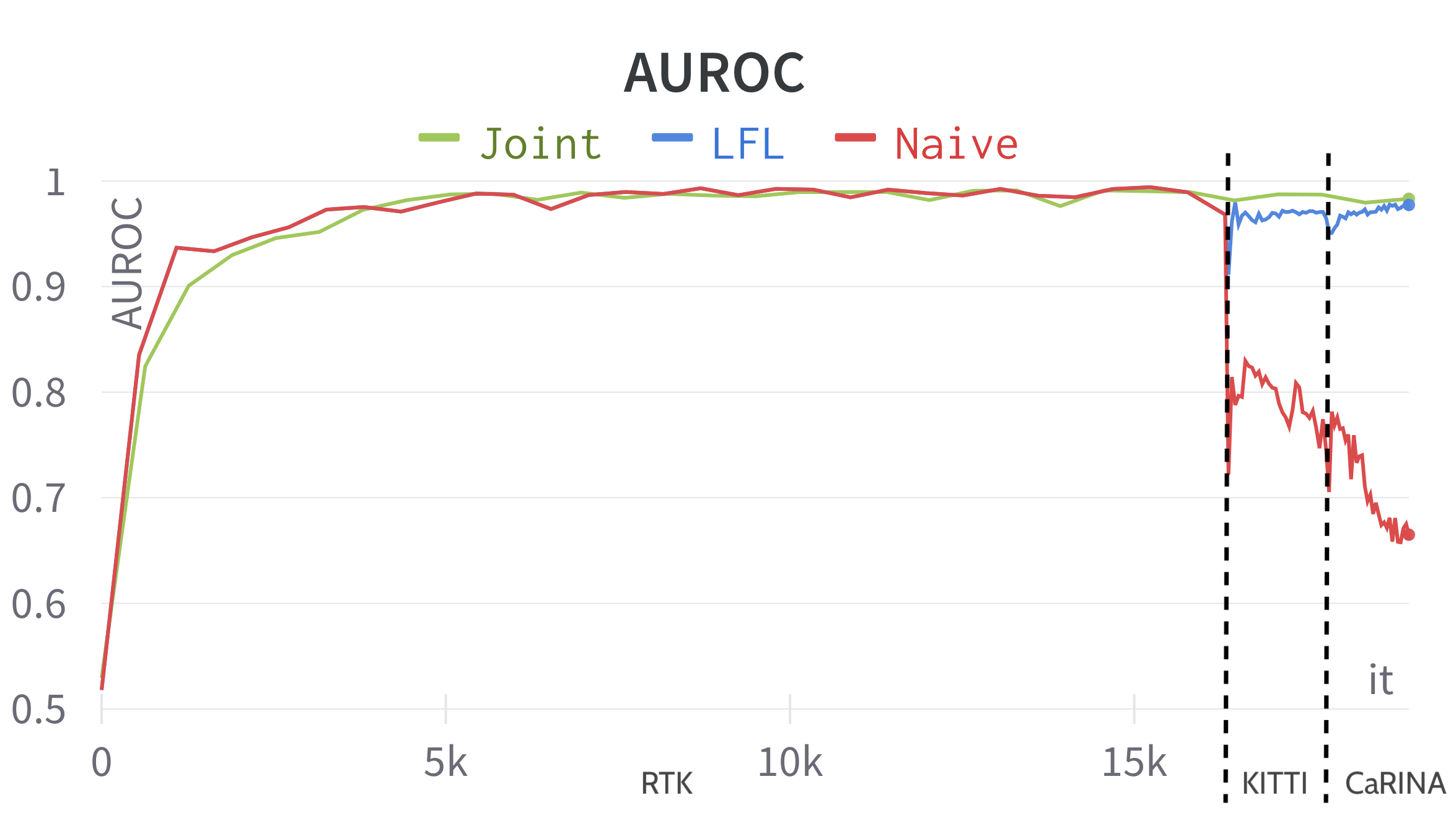}%
    }
    \hfill
    \subfloat{%
        \includegraphics[width=\columnwidth]{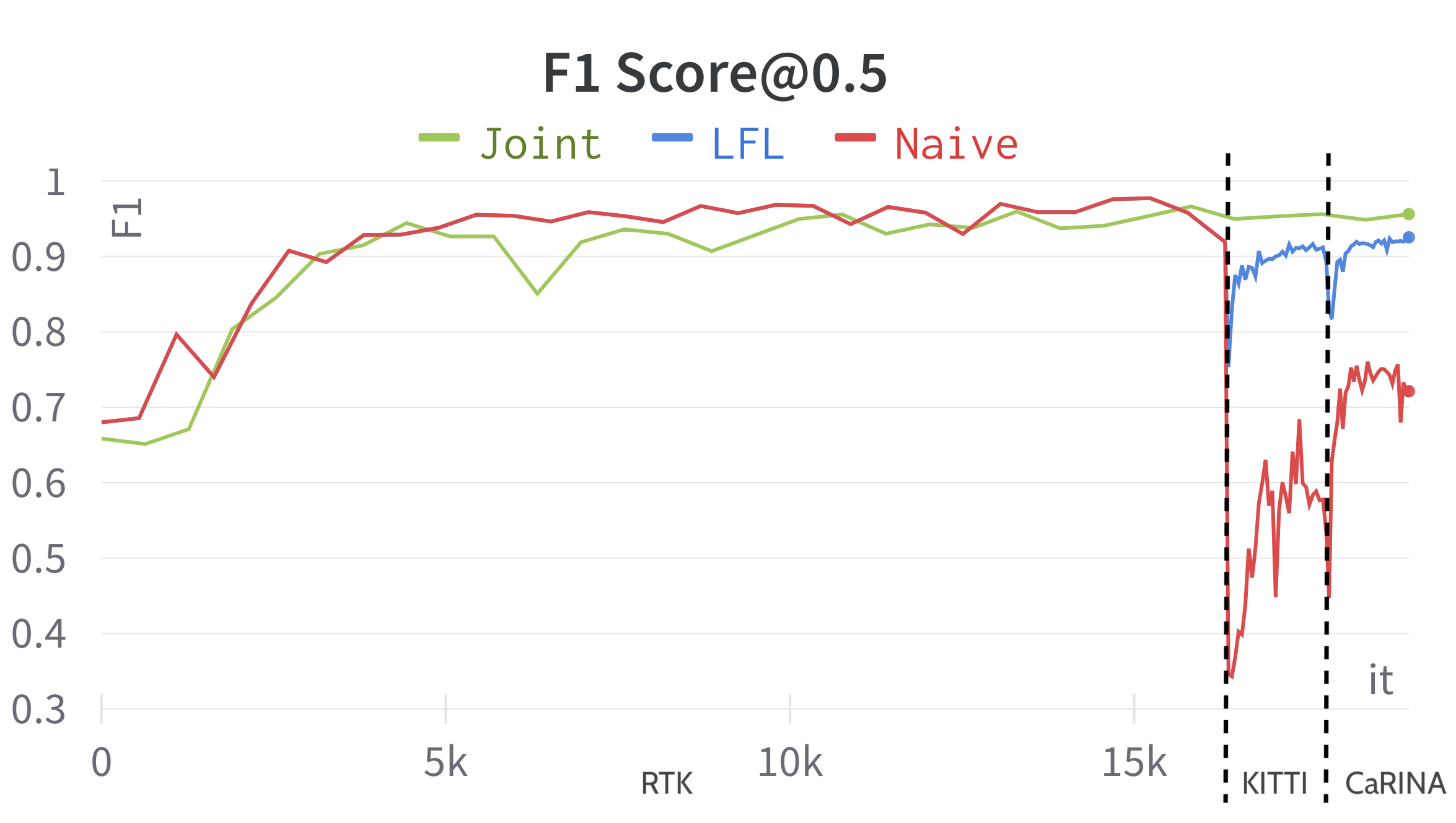}%
    }
    \caption{Trend of AUROC and F1 metrics while training on successive datasets (RTK, followed by KITTI, followed by CaRINA). The metrics are computed over the union of the test sets of these three datasets. It is evident how the Naive strategy sustains a substantial drop in performance when a new dataset is introduced, and never recovers from this occurrence. The continual strategy, LFL, instead, although slightly affected by the change of distribution, quickly stabilizes and maintains almost perfect performances (with respect to a Joint full retraining).}%
    \label{fig:auroc}%
\end{figure}

Incidentally, we notice from \Cref{tab:res_comparison} that, on the last dataset (CaRINA), Naive and LFL strategies achieve slightly higher performance than Joint. Intuitively, this is due to how each strategy deals with the stability-plasticity tradeoff. Naive and LFL can focus their optimization only on new data, thus fitting it slightly better than Joint, which is instead fed random samples from all three datasets.

For additional reference, we report the trend during the entire training of the AUROC and F1 metrics computed on the union of the test sets (\cref{fig:auroc}). As mentioned above, the Joint strategy serves as an upper bound on the performance. We notice that, while the Naive strategy accumulates a significant gap as new datasets are introduced, LFL maintains almost perfect performance throughout the entire training.

\section{CONCLUSION}
\label{sec:conclusion}

In this work, we addressed the problem of cross-dataset adaptation in road surface classification. We showed how continual learning strategies can be adopted to update a given model when new data becomes available, without the need to store and use past datasets, yet avoiding catastrophic forgetting. The efficacy of continual learning for cross-dataset adaptation is confirmed by our experiments, which also highlighted the occurrence of strong forgetting when adopting a naive finetuning approach.
While contributing to tackling this open problem in the field of autonomous vehicles, we believe our formulation provides also the basic tools necessary to apply our approach to other AV tasks. Indeed, the need for updating a model with new data is particularly important for improving the robustness of perception systems when deployed in the real world. With this in mind, we encourage the adoption of continual learning techniques not only in future research, but also in industrial settings, in order to save the computational, economic, and logistic resources due to unnecessary large-scale retraining.

\bibliographystyle{IEEEtran}
\bibliography{references}

\end{document}